\documentclass[lettersize,journal]{IEEEtran}
\usepackage{amsmath,amsfonts}
\usepackage{algorithmic}
\usepackage{algorithm}
\usepackage{array}
\usepackage{textcomp}
\usepackage{stfloats}
\usepackage{verbatim}
\usepackage{graphicx}
\usepackage{cite}
\usepackage{soul}
\usepackage{balance}
\usepackage{makecell}
\usepackage{xcolor}
\usepackage{multirow}
\usepackage{booktabs}
\usepackage{colortbl}
\usepackage{tabularx}
\usepackage{url}
\usepackage{enumerate}
\usepackage{enumitem}
\usepackage[most]{tcolorbox}

\usepackage{listings}
\usepackage{xcolor}

\lstdefinestyle{pcs}{
    language=Python,
    basicstyle=\ttfamily\footnotesize,
    breaklines=true,
    breakatwhitespace=true,
    frame=single,
    numbers=none,
    showstringspaces=false,
    tabsize=2,
    keywordstyle=\color{blue},
    commentstyle=\color{gray},
    stringstyle=\color{teal},
    xleftmargin=0pt,
    xrightmargin=0pt,
    captionpos=b,               
    belowcaptionskip=6pt        
}

\definecolor{sentimentLight}{RGB}{230, 242, 255}
\definecolor{sentimentMedium}{RGB}{179, 214, 255}
\definecolor{sentimentDark}{RGB}{128, 187, 255}

\definecolor{gossipLight}{RGB}{230, 255, 242}
\definecolor{gossipMedium}{RGB}{179, 255, 214}
\definecolor{gossipDark}{RGB}{128, 255, 187}

\definecolor{politiLight}{RGB}{240, 210, 227}
\definecolor{politiMedium}{RGB}{243, 178, 215}
\definecolor{politiDark}{RGB}{242, 132, 194}

\usepackage{balance}
\begin{document}
\balance
\title{PCS: \textbf{P}erceived \textbf{C}onfidence \textbf{S}coring of Black Box LLMs with Metamorphic Relations}

\author{Sina Salimian, Gias Uddin, Shaina Raza, and Henry Leung
\thanks{S. Salimian and H. Leung are with the University of Calgary, S. Raza is with the Vector Institute, and G. Uddin is with York University, Canada. Corresponding author is Gias Uddin. Email: guddin@yorku.ca
}}

\markboth{IEEE Transactions on Software Engieering}%
{Shell \MakeLowercase{\textit{et al.}}: A Sample Article Using IEEEtran.cls for IEEE Journals}


\maketitle

\begin{abstract}
Zero-shot LLMs are now also used for textual classification tasks, e.g., sentiment, bias detection in a sentence/article, etc. 
However, their performance can be suboptimal in such data annotation tasks. We introduce a novel technique that evaluates an LLM's confidence for classifying a textual input by leveraging Metamorphic Relations (MRs). The MRs generate semantically equivalent yet textually divergent versions of the input. Following the principles of Metamorphic Testing (MT), the mutated versions are expected to have annotation labels similar to the input. By analyzing the consistency of an LLM's responses across these variations, we compute a perceived confidence score (PCS) based on the frequency of the predicted labels. PCS can be used for both single and multiple LLM settings (e.g., when multiple LLMs are vetted in a majority voting setup). 
Empirical evaluation shows that our PCS based approach improves the performance of zero-shot LLMs by 9.3\% in textual classification tasks. When multiple LLMs are used in a majority voting setup, we got a performance boost by 5.8\% with PCS.
\end{abstract}


\section{Introduction}

Large Language Models (LLMs) have achieved remarkable success in a wide range of natural language processing (NLP) tasks, including sentiment analysis \cite{zhang-etal-2024-sentiment}, fake news detection \cite{golob2024fact}, and bias identification \cite{kumar2024decoding}. Their performance often rivals that of human annotators \cite{Gligoric2024CanUL}, making them attractive for real-world applications. However, despite their capabilities, LLMs can become unreliable under uncertainty, particularly when internal confidence scores are inaccessible or poorly calibrated \cite{liu2024examining}. This shortcoming has led to growing concerns around the dependability of LLM-generated annotations, especially in sensitive domains requiring transparency and accountability \cite{castelvecchi2021can, jobin2019global}.

Recent work \cite{geng-etal-2024-survey, luo2024understanding} underscores the importance of confidence assessment and interpretability as key pillars for increasing trust in LLMs. Yet, current techniques, such as calibration using tailored datasets or generating adversarial examples \cite{pedapati2024large}, face limitations in flexibility, scalability, and generalizability. Furthermore, traditional ensemble strategies like majority voting overlook model-specific uncertainty and linguistic subtleties, reducing their effectiveness in real-world deployments \cite{wang2019superglue}.

To address these challenges, we introduce a technique for zero-shot LLMs called \textit{Perceived Confidence Score (PCS)}. PCS is a novel method to evaluate the confidence of LLMs in classification tasks by measuring the consistency of their labels across multiple, semantically equivalent variations of the input. These variations are generated using \textit{Metamorphic Relations (MRs)} \cite{10336270, chen2021metamorphic}, which apply controlled, meaning-preserving transformations such as active/passive voice changes, synonym substitutions, and double negation. By leveraging this structured variation, PCS enables robust confidence estimation by learning optimal weights for LLM outputs. We empirically determine the importance of each MR while computing the final label. For example, given $N$ labels from an LLM for a text and its MR variants, we assign a weight to each label using a linear regression model trained on labeled data. The final label is then computed as a weighted combination of these predictions.

We evaluated PCS-based text classification in two settings, single LLM and multiple LLMs. Experiments on three diverse datasets show that PCS significantly outperforms standard methods such as majority voting in classification tasks. For instance, in multiclass sentiment analysis, PCS boosts the AUROC of the Meta-Llama-3-8B-Instruct model by 20.6\%, and the Gemma-2-9B-it model by 2.4\%. On the Gossipcop dataset, PCS improves AUROC by 11.1\% for the Gemma model and by 16.1\% for the Mistral-7B-Instruct-v0.3 model. These results underscore the practical value of PCS in making LLM predictions more trustworthy and reliable.

We focus on confidence estimation in black-box, zero-shot classification settings, where LLM internals (e.g., logits, activations) are inaccessible. We aimed for a model-agnostic solution that avoids self-reports and instead relies on external signals derived from output consistency across semantically equivalent inputs. Our contributions are as follows:

\begin{enumerate}[itemsep=0pt]
    \item \textbf{Metamorphic Relation (MR) Rules}: A set of controlled, semantic-preserving transformations that create diverse textual inputs for robustness testing.
    
    \item \textbf{Perceived Confidence Score (PCS) Framework}: A model-agnostic method for confidence estimation using consistency across LLM responses to varied inputs.
    
    \item \textbf{Comprehensive Evaluations}: Empirical results across multiple datasets demonstrate that PCS improves both confidence assessment and classification performance compared to traditional baselines. To support reproducibility and facilitate future research, we released the replication package containing all code, data, and results at: 
\url{https://anonymous.4open.science/r/PCSA}
\end{enumerate}

\section{Related Work}

\begin{table*}[h]
\centering
\caption{Comparison of existing uncertainty estimation methods organized by category and how PCS addresses their limitations. Each method is listed individually with its specific shortcomings and corresponding PCS advantages.}
\label{tab:method_comparison}
\begin{tabular}{p{2.5cm}p{4.5cm}|p{2.5cm}|p{3cm}|p{3cm}}
\toprule
\textbf{Method Category} & \textbf{Method} & \textbf{Sub-categories} & \textbf{Shortcomings} & \textbf{How PCS Addresses} \\
\midrule
\multirow{5}{2.5cm}{\textbf{Self-Report Methods}} & 
Using CoT prompt variations and asking LLM to assess its own label (correct / incorrect / unsure) \cite{chen2023quantifying}. & 
Self-Report, Consistency-Based, Black-Box & 
Self-report dependency; prompt sensitivity. & 
\multirow{5}{3cm}{PCS avoids reliance on self-confidence reports, vague language, or multi-step prompts. Instead, it uses a single-pass, model-agnostic consistency check via Metamorphic Relations (MRs) to externally and reliably estimate confidence.} \\
\cmidrule{2-4}
& 
Prompting the LLM in a single step to produce top-k guesses and corresponding confidence scores \cite{tian2023just,lin2022teaching}. & 
Self-Report, Black-Box & 
Exaggerated confidence; no external validation. & \\
\cmidrule{2-4}
& 
Using two prompts: one for top-k guesses, and one for their confidence scores \cite{tian2023just}. & 
Self-Report, Black-Box & 
Multi-step prompting; belief-limited. & \\
\cmidrule{2-4}
& 
Using linguistic confidence terms (e.g., "likely", "unlikely") and mapping them to numeric scores \cite{tian2023just}. & 
Self-Report, Black-Box & 
Vague linguistic expressions. & \\
\cmidrule{2-4}
& 
Asking the model to produce the probability that its answer is true (e.g., "P(Answer is True)") \cite{Kadavath}. & 
Self-Report, Black-Box & 
Ungrounded self-assessment; overconfidence. & \\
\midrule
\multirow{5}{2.5cm}{\textbf{White-Box Methods}} & 
Training a calibration model using the model’s internal prediction scores to estimate human-aligned confidence \cite{virk}. & 
White-Box, Supervised, Calibration-Based & 
Requires logit access; open-source only. & 
\multirow{5}{3cm}{PCS requires no internal access (e.g., logits, activations, dropout), avoids complex entropy or clustering, and works with final outputs only—making it efficient, generalizable, and fully compatible with black-box LLM APIs.} \\
\cmidrule{2-4}
& 
Measuring semantic entropy using paraphrases and internal token probabilities \cite{kuhn2023semantic}. & 
White-Box, Consistency-Based & 
Needs internal probabilities; computationally heavy. & \\
\cmidrule{2-4}
& 
Using hidden layer activations to train a supervised classifier that predicts correctness \cite{liu2024uncertainty}. & 
White-Box, Supervised & 
Requires full model internals access. & \\
\cmidrule{2-4}
& 
Clustering semantically similar outputs and computing von Neumann entropy for uncertainty \cite{nikitin2024kernel}. & 
White-Box, Consistency-Based & 
Complex clustering; task-specific generalization. & \\
\cmidrule{2-4}
& 
Using Monte Carlo Dropout during inference to generate multiple outputs and estimate uncertainty from variability \cite{gal2016dropout,shelmanov2021certain,gambardella2024language}. & 
White-Box & 
Requires dropout modification; API-incompatible. & \\
\midrule
\textbf{Perturbation Methods} & 
Using prompt perturbations (paraphrasing, dummy tokens, temperature, and system message changes) and computing confidence based on weighted output similarity \cite{gao2024spuq}. & 
Perturbation, Consistency-Based, Black-Box & 
Uses single LLM; similarity-based weighting; lacks multi-model consensus. & 
PCS uses multiple LLMs, learns weights via regression, and relies on label agreement for external validation. \\
\bottomrule
\end{tabular}
\end{table*}

\subsection{LLMs as Evaluators}

Large Language Models (LLMs) are increasingly being explored as evaluators due to their ability to perform high-level reasoning, subjective judgment, and complex comparisons. Several studies have highlighted the opportunities and risks in using LLMs for evaluation, particularly under ambiguity or domain-specific constraints. For instance, \cite{laskar2024survey} and \cite{saxon2024framework} discuss challenges in using LLMs for nuanced tasks, while \cite{jin2024gpt4v} exposes critical flaws in GPT-4V’s evaluation of medical scenarios. Bias and alignment remain core concerns, as evidenced by works like \cite{thakur2024judgingjudgesevaluatingalignment, 10.5555/3666122.3668142}, which compare LLM judgments against human benchmarks. Other efforts explore how LLMs assess or rank alternatives in zero-shot or few-shot settings \cite{dong2024llmpersonalizedjudge, liusie2024llmcomparativeassessmentzeroshot}, revealing variability in reliability based on context and prompt design.

\subsection{LLMs for Annotation Tasks}

LLMs are also being leveraged for a variety of annotation tasks, offering scalable and low-cost alternatives to traditional human labeling. However, this comes with challenges such as bias, inconsistency, and sensitivity to prompt design. \cite{bhat-varma-2023-large} reported uneven performance when applying LLMs to underrepresented Indic languages. A meta-analysis by \cite{pavlovic-poesio-2024-effectiveness} found that while LLMs are effective and efficient for annotation, their performance can vary significantly depending on domain and task complexity.

Recent techniques aim to improve annotation robustness using few-shot learning or social context cues. Notably, the DAFND framework \cite{liu2024dafnd} and ``Prompt-and-Align'' method \cite{wu2023promptalign} are designed for low-resource settings. Efforts like \cite{tamang2024evaluating} explore using LLM-generated labels to bootstrap more adaptive models. Meanwhile, ethical concerns continue to gain attention, particularly regarding the propagation of stereotypes and misuse of LLM annotations \cite{fan2024subjective, bender2021parrots}.

In the context of misinformation and fake news detection, approaches like FSKD \cite{springer2023fskd} and consistency-based frameworks \cite{nature2024consistency} demonstrate how LLMs can be fine-tuned or guided for higher accuracy while reducing annotation costs. These works highlight the importance of balancing automation with interpretability and reliability in annotation workflows.

\subsection{Metamorphic Testing for LLMs}

Metamorphic Testing (MT) has emerged as a practical and powerful tool for evaluating the reliability of LLMs under controlled perturbations. By applying Metamorphic Relations (MRs), transformations that preserve semantic meaning but alter form, researchers can test the robustness of models without needing labeled examples for every variation. 

\cite{hyun2024metal} applies MT across classification and summarization tasks, illustrating how carefully designed MRs can reveal inconsistencies in model behavior. \cite{xue2024exploring} uses MT in code generation and bug-fixing, while \cite{li2024drowzee} applies MT to detect fact-conflicting hallucinations, increasing the breadth and quality of test data. Other studies \cite{ma2020metamorphic, ribeiro2020beyond, chen2021metamorphic} have also laid foundational work in behavioral testing and test suite generation for NLP models, motivating the application of MRs to LLM evaluations.

Our work builds on these insights, using MRs not only to assess robustness, but also as a signal for optimizing confidence estimation in classification. The ability to measure label consistency across semantically equivalent variants allows us to evaluate LLM reliability without relying on their internal probability estimates.

\subsection{Confidence Assessment for LLM-Based Classifications}

As LLMs are increasingly used in decision-critical contexts, accurately assessing their confidence is vital. Existing techniques can be broadly categorized into self-report methods, white-box methods, and perturbation-based methods (see Table~\ref{tab:method_comparison}).

\textbf{Self-report methods} include approaches where models reflect on their own correctness, such as using CoT variations with self-assessed labels \cite{chen2023quantifying}, generating top-k answers with confidence in single or multiple prompts \cite{tian2023just,lin2022teaching}, or expressing likelihood using linguistic phrases or direct probability scores \cite{Kadavath}. These methods are limited by prompt sensitivity, overconfidence, and a lack of external validation.

\textbf{White-box methods} rely on internal model signals such as logits \cite{virk}, token probabilities \cite{kuhn2023semantic}, hidden activations \cite{liu2024uncertainty}, output clustering \cite{nikitin2024kernel}, or dropout-based sampling \cite{gal2016dropout,shelmanov2021certain,gambardella2024language}. These techniques often require access to proprietary internals, limiting scalability across models and tasks.

\textbf{Perturbation-based methods}, such as SPUQ \cite{gao2024spuq}, modify prompts slightly and measure response stability, but typically use a single model and similarity-weighted scoring, lacking cross-model validation.

Some studies advocate for hybrid frameworks that involve human oversight to mitigate overconfidence and improve reliability \cite{kim2024collaborative}. Others highlight the limitations of current techniques when applied to subjective or morally complex tasks \cite{when_to_make_exceptions_exploring_language_models_as_accounts_of_human__moral_judgment_2022}.

Our PCS framework focuses exclusively on black-box confidence estimation methods that rely solely on model outputs without access to internal activations, logits, or probability distributions. This reflects realistic deployment scenarios where only API-level access is available. Accordingly, we evaluate PCS against recent black-box baselines—including prompt-based self-report methods (e.g., 1S top-k, 2S top-k, Ling 1S) and perturbation-based approaches (e.g., SPUQ)—while excluding white-box methods that require architectural changes or internal model access, which are out of our scope.

\section{Perceived Confidence Scoring (PCS) for Classifications using Metamorphic Relations}\label{pcs-methodology}

\begin{figure}[t]
\vskip 0.2in
\begin{center}
{
\includegraphics[width=0.5\textwidth]{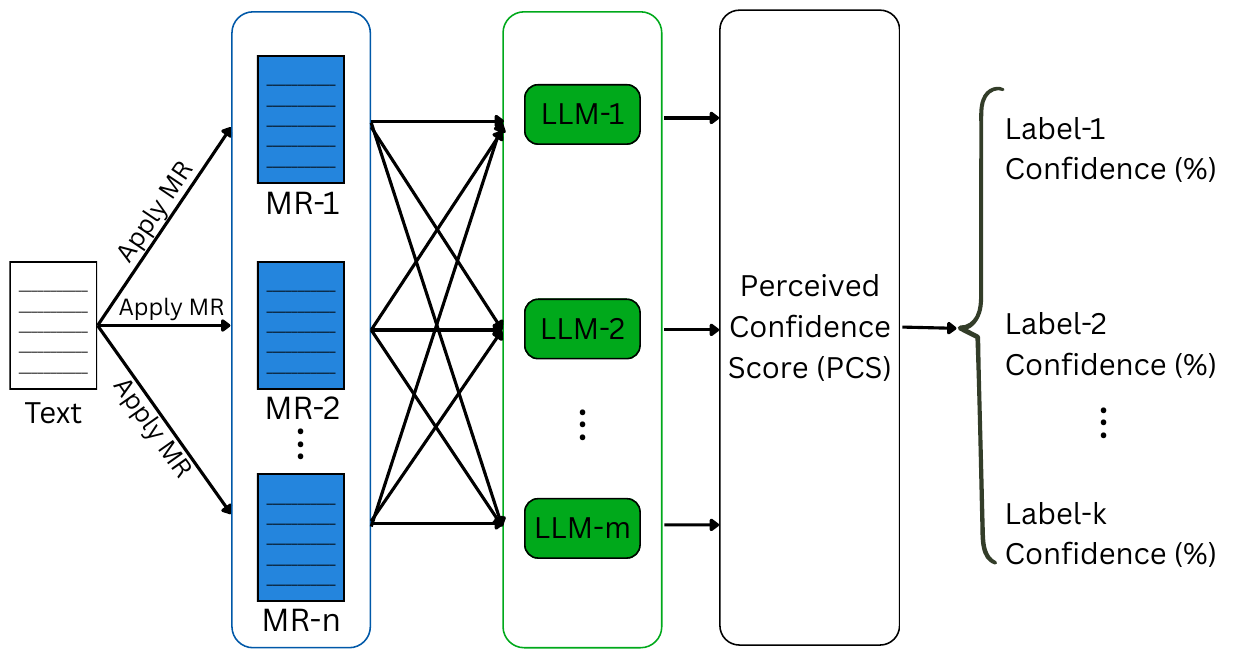}

\caption{Overview of the classification pipeline using Perceived Confidence Score  (PCS) Annotator.}
\label{fig:pipeline}}
\end{center}
\end{figure}

We introduce the \textit{Perceived Confidence Score (PCS)}, a framework for estimating the confidence of Large Language Models (LLMs) in zero-shot classification tasks. PCS is designed for black-box scenarios, where access to model internals, such as logits or attention weights, is not available, and only the final output can be observed.

PCS quantifies confidence based on prediction consistency across semantically equivalent input variations. These variants are generated using \textit{Metamorphic Relations (MRs)}, controlled, meaning-preserving transformations such as synonym substitution, passive-active voice conversion, and double negation. For instance, the sentence \textit{The researcher conducted the experiment} may be rephrased as \textit{The experiment was conducted by the researcher}, and \textit{The policy was criticized by the public} may become \textit{The policy was condemned by the public}. 

If an LLM’s predictions remain stable across such transformations, PCS assigns a higher confidence score; inconsistency implies uncertainty.

To compute PCS, we first learn weights for each LLM and MR based on a labeled training set, using linear regression. At inference time, for any new input, we generate its MR-based variants, obtain predictions from one or more LLMs, and aggregate these responses using the learned weights. The result is a confidence score per class label that reflects not just the model’s prediction, but how consistently it behaves across semantically identical inputs.

PCS avoids reliance on internal scores or self-assessed probabilities. It offers a lightweight, interpretable, and modular approach that is well-suited for deployment in real-world black-box classification pipelines.

\subsection{The Metamorphic Relations (MR)}
\label{MR}

For any given annotation or classification task, the input to our system is a sentence or document requiring a label (e.g., sentiment, veracity, category). To evaluate the consistency and robustness of a model's predictions, we generate multiple semantically equivalent variations of this input using three carefully designed \textit{Metamorphic Relations (MRs)}. These MRs are intentionally task-agnostic and designed to apply broadly across classification tasks without requiring task-specific heuristics. This design choice enhances the generalizability of our framework.

The three MRs used in this study are:
(1) \textbf{MR1 – Active to Passive Voice Transformation}, 
(2) \textbf{MR2 – Double Negation Transformation}, and 
(3) \textbf{MR3 – Synonym Replacement.}

\noindent Each of these transformations is designed to alter the surface form of the text while preserving its core semantic meaning. This allows us to assess whether models make predictions based on genuine understanding rather than superficial lexical or syntactic cues.

\medskip
\noindent\textbf{MR1: Active to Passive Voice Transformation.}  
This MR alters the grammatical structure of a sentence by converting it between active and passive voice. For example, the active sentence \textit{``The Federal Trade Commission (FTC) and 17 state attorneys general have sued Amazon''} becomes \textit{``Amazon has been sued by The Federal Trade Commission (FTC) and 17 state attorneys general''} in passive voice. The transformation leaves the core meaning intact while modifying the syntactic structure, testing the model's resilience to voice shifts.

\medskip
\noindent\textbf{MR2: Double Negation Transformation.}  
This transformation introduces a double negative construction to rephrase a positive or negative statement without altering its semantic intent, i.e., converting a positive phrase into two negative phrases. For instance, \textit{``The White House strongly criticized the US Supreme Court''} is rephrased as \textit{``The White House didn’t weakly criticize the US Supreme Court''}. Although the construction is more complex, the intended meaning remains the same, enabling us to probe the model’s robustness to logical reformulations.

\medskip
\noindent\textbf{MR3: Synonym Replacement.}  
This MR replaces key terms with synonyms to create alternative phrasings while maintaining the original intent. For example, \textit{``The White House strongly criticized the US Supreme Court on Tuesday''} might be rephrased as \textit{``The White House strongly condemned the nation's highest court on Tuesday''}. This form of lexical variation challenges the model to base its decisions on meaning rather than word overlap, as robust models should assign the same label to both phrasings.

\medskip
\noindent Table~\ref{tab:articleMRExamples} illustrates examples of how Metamorphic Relations are applied in practice. The PCS framework is designed to be flexible and extensible: it supports both the integration of new MRs and the combination of multiple transformations.

\subsection {Confidence Scoring of LLMs using MRs}
\label{PCS}
 
For each input instance, we obtain four classification outputs from the LLM: one corresponding to the original input and three additional outputs generated from semantically equivalent variants created through different Metamorphic Relations (MRs). These transformed inputs allow us to assess the stability of the model's prediction in response to controlled linguistic perturbations.

Consider a binary classification task where the LLM is asked to determine whether a news article is \textit{fake} or \textit{real}. After querying the model with the original input and its three MR-based variations, suppose the model returns the label \textit{fake} for three of the inputs and \textit{real} for one. We use this distribution to compute the \textit{Perceived Confidence Score (PCS)} for each class label.

In this example, the PCS for the \textit{fake} label would be $\frac{3}{4}$ (i.e., 75\%), and for the \textit{real} label, it would be $\frac{1}{4}$ (i.e., 25\%). These scores reflect the model’s relative certainty across multiple semantically consistent inputs and form the foundation for making a final, more robust classification decision.

\subsection {PCS across Multiple LLMs}
\label{PCS-multi}

The concept of the Perceived Confidence Score (PCS) is inherently flexible and can be applied not only to a single LLM, but also extended across multiple LLMs operating in parallel. This multi-model setup enhances the reliability of confidence estimation by capturing a broader range of reasoning styles and linguistic behaviors.

Returning to the binary classification task of \textit{fake} vs. \textit{real} described in Section~\ref{PCS}, consider a scenario where we query three distinct LLMs. If each LLM is presented with four input variants (the original plus three MR-based mutations), we obtain a total of 12 predictions for the same input. These outputs collectively represent a richer set of signals from which to infer confidence.

We then compute the PCS for each label by measuring the proportion of predictions that support a given class by each LLM. This aggregation across both textual variations and multiple LLMs enables a more robust and interpretable confidence score, mitigating inconsistencies from any single model or phrasing.



\begin{table*}[t]
\caption{Examples of applying Metamorphic Relations (MRs) to article-based inputs. MR1: Passive/Active Rewrite, MR2: Double Negation transformation, MR3: Synonym Replacement.}
\label{tab:articleMRExamples}
\centering
\resizebox{\textwidth}{!}{%
\begin{tabular}{p{4.2cm}|p{4.2cm}|p{4.2cm}|p{4.2cm}}
\hline
\textbf{Original Article} & \textbf{MR1 (Passive/Active Rewrite)} & \textbf{MR2 (Double Negation)} & \textbf{MR3 (Synonym Replacement)} \\
\hline
The Federal Trade Commission (FTC) and 17 state attorneys general have sued Amazon, alleging the e-commerce behemoth uses its position in the marketplace to inflate prices on other platforms, overcharge sellers and stifle competition. & 
\textcolor{purple}{Amazon have been sued by the Federal Trade Commission (FTC) and 17 state attorneys general}, alleging that it uses its position in the marketplace to inflate prices on other platforms, overcharge sellers and stifle competition. &
The Federal Trade Commission (FTC) and 17 state attorneys general \textcolor{purple}{haven't failed to sue} Amazon, alleging the e-commerce behemoth \textcolor{purple}{doesn't fail to use} its position in the marketplace to inflate prices on other platforms, \textcolor{purple}{doesn't undercharge} sellers and \textcolor{purple}{doesn't fail to stifle} competition. & 
The Federal Trade Commission (FTC) and 17 state attorneys general have \textcolor{purple}{taken legal action against} Amazon, \textcolor{purple}{accusing} the e-commerce \textcolor{purple}{giant} of \textcolor{purple}{exploiting its dominance} in the market to \textcolor{purple}{artificially boost} prices on other platforms, overcharge \textcolor{purple}{vendors} and \textcolor{purple}{suppress rivalry}. \\
\hline
The White House strongly criticised the US Supreme Court on Tuesday for allowing what it called “harmful and unconstitutional” Texas immigration law to go into effect. & 
\textcolor{purple}{The US Supreme Court was strongly criticised by the White House} on Tuesday for allowing what it called a “harmful and unconstitutional” Texas immigration law to go into effect. &
The White House \textcolor{purple}{didn't weakly criticise} the US Supreme Court on Tuesday for \textcolor{purple}{not preventing} what it called a harmful and unconstitutional Texas immigration law from going into effect. & 
The White House strongly \textcolor{purple}{condemned} the \textcolor{purple}{nation's highest court} on Tuesday for \textcolor{purple}{permitting} what it \textcolor{purple}{termed} a harmful and unconstitutional Texas immigration \textcolor{purple}{statute} to \textcolor{purple}{take effect.} \\
\hline
\end{tabular}
}
\end{table*}

\subsection{Zero-Shot LLM Classification with PCS}

The PCS framework enables zero-shot classification by aggregating consistent predictions from multiple LLMs across perturbed versions of a given input. Assume we have access to $N$ zero-shot LLMs, where $N \geq 1$, available for a given classification task. For each input instance, we generate semantically equivalent variants using predefined Metamorphic Relations (MRs). Each LLM is then prompted to assign labels to these inputs, once for the original and once for each of the three mutated versions, resulting in four predictions per model.

To formalize this, let each LLM be denoted by $llm$, and let $I$ represent the set of input variants for a given base question, including the original and its MR-based transformations. Suppose there are $m$ possible classification labels (e.g., $m = 3$ for sentiment analysis with \textit{positive}, \textit{neutral}, \textit{negative}; or $m = 2$ for binary tasks). Let $lbl$ denote a specific label of interest.

We compute the PCS score of label $lbl$ for each LLM by taking a confidence-weighted average over the input variants. Each input is weighted according to a reliability score (e.g., based on the trustworthiness of the MR used), which may be derived empirically during training. This is formalized in Equation~\ref{eq:pcslblllm}:

\begin{equation}
    \label{eq:pcslblllm}
    \text{PCS}_{lbl, llm} = 
    \frac{
        \sum\limits_{I}
        \left( I_{\text{weight}} \cdot 
        \mathbf{1}[I_{\text{label}} = lbl] 
        \right)
    }{
        \sum\limits_{I} I_{\text{weight}}
    }
\end{equation}

\noindent Here, \( I \) refers to each input variant for a given base example, \( I_{\text{label}} \) is the label assigned by the LLM, and \( I_{\text{weight}} \in [0, 1] \) is the confidence weight assigned to that input variant. This formulation computes the confidence-weighted proportion of times the LLM assigns the label \( lbl \), emphasizing input variants deemed more reliable during training.

Once we compute the per-model PCS scores, we aggregate them across all LLMs to obtain a final PCS score for each label. This second layer of weighting considers how much we trust each LLM based on its past performance. Equation~\ref{eq:pcslbl} defines the aggregated PCS for label $lbl$ across all models:

\begin{equation}
    \label{eq:pcslbl}
    \text{PCS}_{lbl} = 
    \frac{
        \sum\limits_{llm} 
        \left( \text{LLM}_{\text{weight}} \cdot \text{PCS}_{lbl, llm} \right)
    }{
        \sum\limits_{llm} \text{LLM}_{\text{weight}}
    }
\end{equation}

\noindent In this equation, \( \text{LLM}_{\text{weight}} \in [0, 1] \) denotes the trust score assigned to each LLM. The final PCS score for label \( lbl \) is computed as the weighted average of scores from individual LLMs. In a sentiment classification task with three possible labels, the output would be three PCS values, each indicating the ensemble’s confidence in a given label assignment. These can then be compared against threshold values to make the final classification decision.

\subsection{Computing Optimal Weights for MRs in PCS}

To produce meaningful PCS scores that reflect both model reliability and transformation robustness, we learn the optimal weights used in Equations~\ref{eq:pcslblllm} and~\ref{eq:pcslbl} using a supervised optimization strategy based on the Linear Regression algorithm. This approach enables the PCS framework to adaptively assign importance to individual LLMs and Metamorphic Relations (MRs) based on empirical performance over a labeled training dataset.



We adopt Linear Regression (LR) to learn the optimal weights because the PCS formulation naturally reduces to a linear regression problem. Specifically, Equations~\ref{eq:pcslblllm} and~\ref{eq:pcslbl} define a linear relationship between the weighted inputs (i.e., the LLM predictions over mutated inputs) and the target classification label. This makes LR a principled and mathematically appropriate choice, as it directly solves the underlying convex optimization problem defined by our framework.

In addition to its theoretical suitability, LR is computationally efficient and produces interpretable weights that offer clear insight into the influence of each LLM and Metamorphic Relation (MR). These characteristics are particularly advantageous in settings where transparency and low-latency retraining are important.

Importantly, the PCS framework is designed to be algorithm-agnostic and flexible. While LR is our default optimizer, the framework can seamlessly accommodate alternative learning strategies. To demonstrate this, we also experimented with a Genetic Algorithm (GA)-based optimization method, which evolves MR and LLM weights through population-based search. The results of these experiments, presented in the supplementary material, show that PCS remains effective under nonlinear optimization strategies as well. This reinforces the robustness of the PCS framework and its adaptability across different optimization paradigms.

The training process is task-specific: users provide a labeled dataset relevant to their classification task of interest (e.g., sentiment analysis, misinformation detection). From this dataset, the algorithm learns how much influence each LLM and each MR should have in the final decision-making process.

\medskip
\noindent The input to the Linear Regression model consists of the following:

\begin{itemize}
    \item \boldmath$X \in \mathbb{R}^{n \times l \times m}$\unboldmath: A 3-dimensional tensor representing the PCS-aligned predictions across the training set, where $n$ is the number of samples, $l$ is the number of LLMs ($N_L$), and $m$ is the number of MRs ($N_M$).
    
    \item \boldmath$Y \in \{c_1, \dots, c_k\}^n$\unboldmath: The ground truth labels associated with each of the $n$ training samples.
    
    \item \boldmath$C = \{c_1, \dots, c_k\}$\unboldmath: The complete set of possible class labels for the task.
\end{itemize}

\noindent The regression algorithm then learns a parameter vector 
\[
\boldsymbol{\theta} = [\mathbf{w_M}, \mathbf{w_L}],
\]
where:

\begin{itemize}
    \item \boldmath$\mathbf{w_M} = [w_{M_1}, \dots, w_{M_{N_M}}]$\unboldmath: A vector of weights representing the relative importance of each MR.
    
    \item \boldmath$\mathbf{w_L} = [w_{L_1}, \dots, w_{L_{N_L}}]$\unboldmath: A vector of weights representing the relative trustworthiness of each LLM.
\end{itemize}

These weights allow PCS to balance the contributions from different LLMs and input transformations, enhancing robustness against unreliable models or misleading perturbations. The optimization process is subject to the following constraints to ensure interpretability and normalized influence:

\begin{align*}
\sum_{i=1}^{N_M} w_{M_i} &= 1,\quad w_{M_i} \in [0, 1] \\
\sum_{i=1}^{N_L} w_{L_i} &= 1,\quad w_{L_i} \in [0, 1]
\end{align*}

\noindent These constraints enforce that the learned weights form valid convex combinations, allowing PCS scores to be interpreted as normalized confidence values. By grounding weight learning in labeled data and applying it to both MRs and LLMs, our method ensures that classification decisions are guided by empirically validated components.

\section{Evaluation}
\label{sec:evaluation}
We evaluate the effectiveness of the PCS framework across two representative classification tasks: (1) binary \textit{real/fake} news detection and (2) multiclass sentiment analysis with labels \textit{positive}, \textit{neutral}, and \textit{negative}. Our goal is to assess PCS's ability to provide reliable confidence estimates, particularly in zero-shot settings where no task-specific fine-tuning is performed.

To ensure a fair and rigorous comparison, we benchmark PCS against several baselines that are explicitly designed to assess or approximate confidence in LLM outputs:

\noindent \textbf{Single-model Zero-shot Predictions}: An individual LLM directly assigns a label to each input without additional perturbations or ensembling.
    
\noindent \textbf{Multi-model Majority Voting}: Predictions from multiple LLMs are aggregated; confidence is inferred based on inter-model agreement.
    
\noindent \textbf{1S top-$k$} and \textbf{2S top-$k$} \cite{tian2023just}: In the 1S (single-step) setting, the LLM produces its top-$k$ label predictions in a single inference, each accompanied by a confidence score. In contrast, the 2S (two-step) setting separates the processes: the first step generates a predicted label, and the second step independently estimates the confidence for that prediction.
    
\noindent \textbf{Ling 1S} \cite{tian2023just}: Instead of numerical scores, the LLM selects from a set of natural language uncertainty expressions (e.g., \textit{Likely}, \textit{Almost no chance}), which are then mapped to calibrated probabilities based on human ratings.
    
\noindent \textbf{SPUQ} \cite{gao2024spuq}: A perturbation-based framework that introduces semantic changes through paraphrasing, dummy tokens, or prompt engineering.

PCS is most closely related to SPUQ, which also uses input perturbations to infer confidence. However, PCS introduces several key differences: it learns optimal weights for each transformation through supervised training, whereas SPUQ assigns weights based on cosine similarity to the original input; it does not rely on the model’s self-reported confidence or probabilities, unlike SPUQ; and it uses semantically grounded transformations derived from Metamorphic Relations (MRs), rather than relying on purely syntactic or token-level perturbations.

These baselines were chosen based on their explicit focus on confidence estimation in LLMs. Each method offers a unique perspective on the problem, ranging from linguistic uncertainty to model agreement, and therefore provides a meaningful point of comparison for PCS in multiclass classification scenarios. We specifically selected recent black-box methods that operate without requiring access to internal model representations (e.g., logits or hidden states), making them directly comparable to PCS. Other white-box methods were excluded as they fall outside the scope of our work, which focuses on confidence estimation in settings where model internals are inaccessible.

To evaluate PCS, we applied three task-agnostic Metamorphic Relations (MRs) to each input to generate semantically equivalent variants. These include voice alternation, synonym substitution, and double negation, as detailed in Section~\ref{MR}. This setup allowed us to conduct a comprehensive analysis along four core research questions:

\begin{itemize}
    \item \textbf{RQ1:} Can PCS outperform individual zero-shot LLMs in classification tasks?
    
    \item \textbf{RQ2:} Can PCS outperform existing confidence estimation techniques?
    
    \item \textbf{RQ3:} Can PCS outperform multi-LLM ensemble baselines such as majority voting?
    
    \item \textbf{RQ4:} How does each Metamorphic Relation contribute to final classification performance?
\end{itemize}

To address \textbf{RQ1}, we directly compared PCS to the outputs of individual LLMs operating on unaltered inputs. For \textbf{RQ2}, we evaluated PCS against other single-model confidence frameworks such as 1S/2S top-$k$, Ling 1S, and SPUQ. In \textbf{RQ3}, we examined how PCS compares to multi-model ensemble methods, particularly majority voting over multiple zero-shot LLMs. Lastly, in \textbf{RQ4}, we conducted a detailed ablation study to evaluate the marginal contribution of each MR, analyzing whether certain transformations disproportionately contribute to performance gains.

This evaluation protocol enables us to holistically assess the accuracy, consistency, and robustness of PCS across both task-specific and model-agnostic baselines.

\subsection{Datasets}

The PCS framework is designed for general use and can be applied to any textual dataset or classification task, regardless of domain, input structure, or label granularity. To demonstrate its versatility and real-world applicability, we evaluate PCS on three publicly available datasets spanning both binary and ternary classification tasks. These datasets are widely used in the academic community for benchmarking, and our use of them is strictly for research purposes without redistribution. Full dataset details are provided in the supplementary material.

\medskip
\noindent\textbf{FakeNewsNet-Gossipcop} \cite{shu2018fakenewsnet, shu2017fake, shu2017exploiting, shu2020fakenewsnet}  
\vspace{-1mm}

This dataset includes 393 news articles labeled as either \textit{real} or \textit{fake}. The labels are manually curated and verified for factual correctness and serve as ground truth for fake news detection tasks. The content in this dataset typically reflects entertainment and celebrity-related stories, providing a challenging testbed for evaluating classification robustness in less formal contexts.

\medskip
\noindent\textbf{FakeNewsNet-Politifact} \cite{shu2018fakenewsnet, shu2017fake, shu2017exploiting, shu2020fakenewsnet}  
\vspace{-1mm}

This dataset comprises 135 political news samples, again annotated with binary \textit{real} or \textit{fake} labels. Each label is manually verified through fact-checking processes on the Politifact platform. The language used in this dataset tends to be more formal and policy-oriented, introducing domain diversity in our evaluation setup.

\medskip
\noindent\textbf{Multiclass-Sentiment-Analysis} \cite{hf_multiclass_sentiment}  
\vspace{-1mm}

Designed for ternary sentiment classification, this dataset includes 405 text samples labeled as \textit{positive}, \textit{negative}, or \textit{neutral}. The annotations are manually reviewed and verified, ensuring high-quality ground truth for assessing LLMs’ sensitivity to nuanced emotional expression. This dataset enables us to test PCS in more subjective and fine-grained classification tasks beyond binary distinctions.

\medskip

These datasets offer a diverse evaluation landscape (covering binary and multiclass tasks, formal and informal language, and objective vs. subjective labels), providing a solid foundation to assess PCS’s robustness and generalizability across contexts.

\subsection{Evaluation Metrics}  

To rigorously evaluate the effectiveness of PCS, we compare its performance against two primary baseline settings that reflect common practices in zero-shot classification with LLMs:

\begin{itemize}
    \item \textbf{(i) Single-Model Setting:} Classification is performed using a single zero-shot LLM, without any input perturbation or ensemble aggregation. This baseline reflects the most basic deployment of LLMs for classification tasks and serves as a lower-bound reference.
    
    \item \textbf{(ii) Multi-Model (Majority Voting) Setting:} Predictions are aggregated from multiple LLMs, and the final label is selected via majority voting, i.e., the label that appears most frequently across models is chosen. 
\end{itemize}

To quantitatively assess performance, we use the Area Under the Receiver Operating Characteristic Curve (AUROC) as our primary evaluation metric. AUROC is particularly suitable in our setting because it offers a threshold-independent assessment of classification quality. It captures the trade-off between true positive and false positive rates across all possible decision thresholds, providing a more holistic view of classifier reliability.

\subsection{Models}

Our evaluation incorporates three state-of-the-art instruction-tuned Large Language Models (LLMs), each representing a distinct architecture and training paradigm: \textbf{LLaMA} (Meta-Llama-3-8B-Instruct), \textbf{Mistral} (Mistral-7B-Instruct-v0.3), and \textbf{Gemma} (gemma-2-9b-it). These models were selected for their strong zero-shot generalization capabilities and widespread adoption in the research community.

\subsection{Setup and Hyperparameters}

To generate semantically consistent input variations, we employed the Metamorphic Relations (MRs) described in Section~\ref{MR}, using them as controlled perturbation mechanisms to evaluate the robustness of LLM-based classification. Specifically, input mutations were produced using the \texttt{LLaMA3.1-8B-Instruct} model, which served as the text mutator across all three transformation types (Prompts are available in the supplementary material). 

All LLMs and text mutation routines were implemented using the Hugging Face \texttt{Transformers} library, which provided a standardized and efficient interface for model loading and inference. Experiments were conducted on an NVIDIA V100 GPU, ensuring fast parallel processing and reproducible performance. Further details on the model configurations are provided in the supplementary material.

\section{Results And Discussion}
\label{sec:results}

\begin{table}[t]
\centering
\caption{Comparing PCS with single zero-shot LLM approach (L: Meta-Llama-3-8B-Instruct, M: Mistral-7B-Instruct-v0.3, and G: gemma-2-9b-it). The last row shows the average performance of PCS compared to the zero-shot LLMs.}
\label{tab:singleLLMs}
{

\begin{tabular}{lccc}
\toprule
\multirow{2}{*}{\textbf{Dataset}} & \multirow{2}{*}{\textbf{LLM}} & \multicolumn{2}{c}{\textbf{AUROC $\boldsymbol{\uparrow}$ (Temp = Default)}} \\
\cmidrule(lr){3-4} 
& & Zero-shot & PCS \\
\midrule
\multirow{3}{*}{\shortstack{Multiclass\\Sentiment\\Analysis}} 
& L & 0.68 & \textbf{0.82} \\
& M & \textbf{0.70} & \textbf{0.70} \\
& G & 0.85 & \textbf{0.87} \\
\midrule
\multirow{3}{*}{Gossipcop} 
& L & 0.54 & \textbf{0.62} \\
& M & 0.62 & \textbf{0.72} \\
& G & 0.72 & \textbf{0.80} \\
\midrule
\multirow{3}{*}{Politifact} 
& L & \textbf{0.88} & \textbf{0.88} \\
& M & 0.73 & \textbf{0.86} \\
& G & 0.85 & \textbf{0.86} \\
\midrule


\multicolumn{2}{c}{\textbf{Average Relative Improvement}} & \multicolumn{2}{c}{\textbf{+9.3\%}} \\
\midrule
\multicolumn{2}{c}{\textbf{Paired t-test p-value}} & \multicolumn{2}{c}{\textbf{0.01}} \\
\bottomrule
\end{tabular}

}

\end{table}





\subsection{Single Zero-Shot LLM Performance (RQ1)}

We begin our evaluation by assessing the effectiveness of PCS when applied to a single LLM performing zero-shot annotation. In the baseline setting, the LLM classifies each input using a single prompt without any form of input augmentation (prompt examples are provided in the supplementary material). In contrast, the PCS approach generates multiple semantically equivalent variations of the input using the Metamorphic Relations (MRs) defined in Section~\ref{MR}. The LLM is then queried with each of these variations ( including the original input) and returns a predicted label for each one.

These predictions are subsequently aggregated using the PCS framework, which assigns optimized weights to each variant. The weights are learned from training data using the Linear Regression algorithm, allowing PCS to emphasize more reliable input forms and de-emphasize noisier ones. The final classification decision is based on the weighted aggregation of all predictions, offering a more robust and confidence-aware alternative to standard zero-shot annotation.

Table~\ref{tab:singleLLMs} presents the AUROC scores across three datasets—\textit{Multiclass Sentiment Analysis}, \textit{Gossipcop}, and \textit{Politifact}—using three individual LLMs: LLaMA (Meta-Llama-3-8B-Instruct), Mistral (Mistral-7B-Instruct-v0.3), and Gemma (gemma-2-9b-it), denoted respectively as L, M, and G. All results use the models' default temperature settings; results for a lower temperature setting (0.1) are available in the supplementary material.

Across all three datasets, PCS notably enhances classification performance. For the \textit{Multiclass Sentiment Analysis} task, the LLaMA model’s AUROC score improves from 0.68 to 0.82, a relative increase of 20.1\%. For \textit{Gossipcop}, the Gemma model sees an increase from 0.72 to 0.80, representing an 11.1\% relative improvement. Similarly, on the \textit{Politifact} dataset, the Mistral model’s AUROC rises from 0.73 to 0.86, reflecting a 17.8\% relative gain. While PCS may not always outperform the zero-shot baseline for every model-dataset pair (e.g., the M model in Sentiment Analysis), it generally provides significant and consistent performance boosts across the board. On average, PCS improves AUROC by 9.3\% over standard single-model zero-shot classification.

To evaluate the significance of PCS’s performance gains, we apply a two-tailed paired t-test on AUROC scores across 9 model--dataset pairs. The test statistic is:

\begin{equation}
t = \frac{\bar{d}}{s_d / \sqrt{n}}
\label{eq:ttest}
\end{equation}

\noindent where $\bar{d}$ is the mean difference, $s_d$ the standard deviation, and $n = 9$. Using \texttt{scipy.stats.ttest\_rel}, we obtain a \textbf{p-value of 0.01} (Table~\ref{tab:singleLLMs}), confirming that PCS’s improvements are statistically significant at the 1\% level.

\medskip
{\centering
\begin{tcolorbox}
[width=1\columnwidth, boxrule=0.5pt, colback=gray!10, arc=4pt,
                  left=6pt, right=6pt, top=6pt, bottom=6pt, boxsep=0pt]
    \textbf{Answer to RQ1:} As shown in Table~\ref{tab:singleLLMs}, incorporating PCS into single zero-shot LLM classification yields a significant average relative improvement of 9.3\% in AUROC across all benchmark datasets.
\end{tcolorbox}
}

\subsection{Comparison with Baseline Confidence Estimation Methods (RQ2)}

\begin{table}[t]
\centering
\caption{Performance comparison using a single LLM (Gemma-2-9b-it): PCS vs. baseline methods with statistical significance. All p-values represent comparisons against the PCS method. Significance levels are denoted as follows: \textcolor{red}{\textbf{***}} $p < 0.001$ (highly significant), \textcolor{red}{\textbf{**}} $p < 0.01$ (very significant), \textcolor{orange}{\textbf{*}} $p < 0.05$ (significant), and \textcolor{gray}{\textbf{n.s.}} $p \geq 0.05$ (not significant).}
\label{tab:baseline}
{

\resizebox{\linewidth}{!}{

\begin{tabular}{llccc}
\toprule
\textbf{Dataset} & \textbf{Method} & \textbf{AUROC \(\boldsymbol{\uparrow}\)} & \textbf{p-value} & \textbf{Significance} \\
\midrule
\multirow{5}{*}{\shortstack{Multiclass\\Sentiment\\Analysis}} 
 & \textbf{PCS (Reference)}     & \textbf{0.87} & -- & -- \\
 & 1S top-2                     & 0.82 & 0.033 & \textcolor{orange}{*} \\
 & 2S top-2                     & 0.81 & 0.022 & \textcolor{orange}{*} \\
 & Ling 1S                      & 0.73 & $<0.001$ & \textcolor{red}{***} \\
 & SPUQ                         & 0.67 & $<0.001$ & \textcolor{red}{***} \\
\cmidrule(lr){1-5}
\multirow{5}{*}{Gossipcop}
 & \textbf{PCS (Reference)}     & \textbf{0.80} & -- & -- \\
 & Ling 1S                      & 0.75 & 0.059 & \textcolor{gray}{ns} \\
 & SPUQ                         & 0.71 & 0.002 & \textcolor{red}{**} \\
 & 1S top-2                     & 0.57 & $<0.001$ & \textcolor{red}{***} \\
 & 2S top-2                     & 0.57 & $<0.001$ & \textcolor{red}{***} \\
\cmidrule(lr){1-5}
\multirow{5}{*}{Politifact}
 & \textbf{PCS (Reference)}     & \textbf{0.86} & -- & -- \\
 & 2S top-2                     & 0.85 & 0.441 & \textcolor{gray}{ns} \\
 & Ling 1S                      & 0.78 & 0.041 & \textcolor{orange}{*} \\
 & SPUQ                         & 0.78 & 0.051 & \textcolor{gray}{ns} \\
 & 1S top-2                     & 0.60 & $<0.001$ & \textcolor{red}{***} \\
\bottomrule
\end{tabular}
}
}
\end{table}

To assess the relative effectiveness of PCS as a confidence estimation framework, we compared its performance against several established baseline methods. Table~\ref{tab:baseline} summarizes the results of this comparison using the same underlying language model, \texttt{Gemma-2-9b-it}, across all approaches to ensure a fair and controlled evaluation. The baselines include: \textbf{1S Top-2}, \textbf{2S Top-2}, \textbf{Ling 1S}, and \textbf{SPUQ} (See Section \ref{sec:evaluation}).

Our evaluation spans three diverse classification tasks—\textit{Multiclass Sentiment Analysis}, \textit{Gossipcop}, and \textit{Politifact}, allowing us to test PCS across both ternary (sentiment) and binary (fake news) classification domains. Across all datasets, PCS consistently achieves the highest AUROC scores: 0.87 for Multiclass Sentiment Analysis, 0.80 for Gossipcop, and 0.86 for Politifact. 

To assess the statistical reliability of PCS’s performance relative to existing methods, we conducted \textit{pairwise significance testing} for each dataset, treating PCS as the reference method. For each baseline, we tested whether the difference in AUROC scores compared to PCS was statistically significant using two-tailed paired t-tests, following the formulation in Equation~\ref{eq:ttest}. The null hypothesis for each comparison was that there is no difference in mean AUROC between PCS and the baseline.

Out of 12 comparisons (4 baselines $\times$ 3 datasets), PCS achieved statistically significant improvements ($p < 0.05$) in \textbf{9 cases}, with several reaching strong significance levels ($p < 0.001$). These results are summarized in Table~\ref{tab:baseline}, with color-coded significance levels: \textcolor{red}{\textbf{***}} ($p < 0.001$), \textcolor{red}{\textbf{**}} ($p < 0.01$), \textcolor{orange}{\textbf{*}} ($p < 0.05$), and \textcolor{gray}{\textbf{ns}} (not significant). Notably, PCS consistently outperformed SPUQ and the 1S/2S top-$k$ variants across most datasets. While the improvement over 2S top-2 on \textit{Politifact} was not statistically significant ($p = 0.441$), and the comparison with SPUQ was marginally non-significant ($p = 0.051$), PCS still achieved the highest AUROC in every dataset.

\medskip
{\centering
\begin{tcolorbox}
[width=1\columnwidth, boxrule=0.5pt, colback=gray!10, arc=4pt,
                  left=6pt, right=6pt, top=6pt, bottom=6pt, boxsep=0pt]
    \textbf{Answer to RQ2:} As shown in Table~\ref{tab:baseline}, PCS consistently outperforms all baseline methods in AUROC across three datasets, achieving statistically significant gains in 9 out of 12 comparisons ($p < 0.05$). These results highlight PCS as a robust and superior framework for confidence assessment in LLM-based classification.
\end{tcolorbox}
}

\subsection{Multi-LLM Zero-Shot Classification (RQ3)}

\begin{table}[t]
\centering

\caption{Performance of the PCS in multi LLM setting (L: Meta-Llama-3-8B-Instruct, M: Mistral-7B-Instruct-v0.3, and G: gemma-2-9b-it). The last row shows the average performance of PCS compared to the MV (Majority Voting) technique.}
\label{tab:MajorityVoting}
{

\begin{tabular}{lclc}
\toprule
\multirow{2}{*}{\textbf{Dataset}} & \multirow{2}{*}{\textbf{LLMs}} & \multicolumn{2}{c}{\textbf{AUROC $\boldsymbol{\uparrow}$}} \\
\cmidrule(lr){3-4}
& & MV & PCS \\
\midrule
\multirow{4}{*}{\shortstack{Multiclass\\Sentiment\\Analysis}} 
& M+G       & \textbf{0.87} & \textbf{0.87} \\
& L+M       & 0.76 & \textbf{0.80} \\
& L+G       & 0.85 & \textbf{0.88} \\
& L+M+G     & 0.86 & \textbf{0.87} \\
\midrule
\multirow{4}{*}{Gossipcop} 
& M+G       & 0.75 & \textbf{0.81} \\
& L+M       & 0.59 & \textbf{0.70} \\
& L+G       & 0.67 & \textbf{0.72} \\
& L+M+G     & 0.69 & \textbf{0.75} \\
\midrule
\multirow{4}{*}{Politifact} 
& M+G       & 0.86 & \textbf{0.91} \\
& L+M       & 0.88 & \textbf{0.91} \\
& L+G       & 0.85 & \textbf{0.89} \\
& L+M+G     & 0.88 & \textbf{0.91} \\
\midrule
\multicolumn{2}{c}{\textbf{Average Relative Improvement}} & \multicolumn{2}{c}{\textbf{+5.8\%}} \\
\midrule
\multicolumn{2}{c}{\textbf{Paired t-test p-value}} & \multicolumn{2}{c}{\textbf{$<0.01$}} \\
\bottomrule
\end{tabular}

}

\end{table}

To address \textbf{RQ3}, we evaluated the performance of PCS in settings involving multiple zero-shot LLMs working in combination. Specifically, we tested all possible pairwise and full combinations of the three models used in our study, \texttt{Meta-Llama-3-8B-Instruct} (L), \texttt{Mistral-7B-Instruct-v0.3} (M), and \texttt{Gemma-2-9b-it} (G). The performance of each configuration was compared against the ensemble technique of Majority Voting (MV), in which confidence for each possible label is assigned based on the proportion of models that predict it.

Table~\ref{tab:MajorityVoting} summarizes the AUROC results for all model combinations using default temperature settings. Additional experiments with lower temperature settings ($T = 0.1$) are reported in the supplementary material to assess the effect of output variability on classification performance.

Our results show that PCS consistently outperforms the Majority Voting baseline across model combinations and datasets. For example, in the \textit{Multiclass Sentiment Analysis} task, the PCS configuration combining L and G achieved an AUROC of 0.88, outperforming both individual models (L = 0.68, G = 0.85) as well as their Majority Voting ensemble (0.85). Similarly, for the \textit{Politifact} dataset, combinations such as M+G, L+M, and L+M+G all achieved an AUROC of 0.91, surpassing both the MV baseline (0.86, 0.88, 0.88, respectively) and the individual model scores (L = 0.88, M = 0.73, G = 0.85; see Table~\ref{tab:singleLLMs}).

On average, PCS yields a 5.8\% relative improvement in AUROC over the Majority Voting approach across all multi-LLM combinations. This consistent gain demonstrates PCS’s ability to intelligently aggregate predictions from diverse models, leveraging their complementary strengths rather than treating them as equal contributors.

\medskip
{\centering
\begin{tcolorbox}
[width=1\columnwidth, boxrule=0.5pt, colback=gray!10, arc=4pt,
                  left=6pt, right=6pt, top=6pt, bottom=6pt, boxsep=0pt]
    \textbf{Answer to RQ3:} As shown in Table~\ref{tab:MajorityVoting}, PCS provides a consistent and significant improvement in AUROC over Majority Voting, with an average relative gain of 5.8\% across multi-LLM zero-shot classification tasks.
\end{tcolorbox}
}

\subsection{Impact of Individual Metamorphic Relations (RQ4)}

To explore the specific contribution of each Metamorphic Relation (MR) to overall classification performance, we conducted an ablation study by applying each MR independently and comparing results to both the unaltered input setting (No MR) and the full PCS ensemble. The results, presented in Table~\ref{tab:MajorityVotingEachMR}, reveal that while individual MRs do not always outperform the baseline (No MR), their combination via PCS consistently yields superior results.

For example, on the \textit{Politifact} dataset using the Mistral (M) model, applying MR3 alone results in a modest AUROC of 0.69, which is lower than the No MR score of 0.73. However, both MR1 and MR2 improve upon the baseline, achieving AUROC scores of 0.77 each. Most notably, when PCS combines all scenarios (i.e., MR1, MR2, MR3, and No MR), the AUROC rises to 0.86, clearly surpassing the performance of any single configuration. This demonstrates that while individual MRs may have variable impact, their collective strength lies in complementary coverage of model behavior.

In some cases, Majority Voting (MV) can reduce performance compared to the best individual LLM. For example, on the \textit{Gossipcop} dataset, Gemma-2-9b-it (G) alone achieves an AUROC of 0.72, while MV drops to 0.69. This drop is likely due to weaker models—LLaMA (L) with 0.54 and Mistral (M) with 0.62—diluting the stronger predictions from G. However, PCS overcomes this issue, improving performance to 0.75 and further to 0.80 when optimized weights prioritize G, demonstrating its ability to leverage individual LLMs' strengths effectively.

Although PCS generally provides robust performance, it does not universally outperform the best-performing individual MR. For example, on the \textit{Gossipcop} dataset with LLaMA (L), MR3 slightly outperforms PCS (0.63 vs. 0.62 AUROC). Similarly, on the \textit{Multiclass Sentiment Analysis} dataset with Mistral (M), the No MR setup performs on par with PCS. However, such cases are rare, and the performance differences are marginal. As Table~\ref{tab:pvalues} shows, these differences are not statistically significant, reaffirming PCS's overall advantage.

The statistical analysis in Table~\ref{tab:pvalues} further reveals that MR1 and MR3 typically perform comparably to the No MR baseline, indicating these perturbations preserve model behavior. In contrast, MR2 often underperforms relative to others, suggesting it contributes less to standalone accuracy. However, MR2 is not without merit: in certain instances, such as with L on Gossipcop and M on Politifact, it performs competitively or even better than other settings. Moreover, MR2 occasionally correctly classifies examples that all other configurations mislabel, highlighting its unique utility in the ensemble.

These performance differences across MRs underscore the sensitivity of LLMs to small changes in linguistic structure. Even minor changes in input structure, while preserving the original meaning, can lead to noticeably different responses from the models.

\medskip
{\centering
\begin{tcolorbox}
[width=1\columnwidth, boxrule=0.5pt, colback=gray!10, arc=4pt,
                  left=6pt, right=6pt, top=6pt, bottom=6pt, boxsep=0pt]
    \textbf{Answer to RQ4:} As shown in Tables~\ref{tab:MajorityVotingEachMR} and~\ref{tab:pvalues}, individual MRs vary in their effectiveness across models and tasks. While no single MR is universally optimal, PCS successfully combines their diverse contributions to consistently outperform any single variant, demonstrating its strength in leveraging controlled semantic perturbations for more reliable classification.
\end{tcolorbox}
}

\begin{table}[t]
\centering%
\caption{Comparing the AUROC performance of zero-shot LLMs across scenarios involving MR1, MR2, MR3, and No MR with the PCS, which utilizes all MRs.}
\label{tab:MajorityVotingEachMR}
{

\resizebox{\linewidth}{!}{
\begin{tabular}{ccccccc}
\toprule
\textbf{Dataset} & \textbf{LLM} & \textbf{No MR} & \textbf{MR1} & \textbf{MR2} & \textbf{MR3} & \textbf{PCS}\\
\midrule
\multirow{4}{*}{\centering \shortstack{Multiclass\\ Sentiment\\Analysis}} 
    & L & \cellcolor{sentimentMedium}0.68 & \cellcolor{sentimentMedium}0.71 & \cellcolor{sentimentLight}0.53 & \cellcolor{sentimentDark}0.75 & \cellcolor{sentimentDark}\textbf{0.82} \\
    & M & \cellcolor{sentimentMedium}\textbf{0.70} & \cellcolor{sentimentMedium}0.62 & \cellcolor{sentimentLight}0.52 & \cellcolor{sentimentMedium}0.61 & \cellcolor{sentimentMedium}\textbf{0.70}\\
    & G & \cellcolor{sentimentDark}0.85 & \cellcolor{sentimentDark}0.77 & \cellcolor{sentimentMedium}0.62 & \cellcolor{sentimentDark}0.74 & \cellcolor{sentimentDark}\textbf{0.87} \\
    & L+M+G (MV) & \cellcolor{sentimentDark}0.86 & \cellcolor{sentimentDark}0.80 & \cellcolor{sentimentMedium}0.58 & \cellcolor{sentimentDark}0.79 & \cellcolor{sentimentDark}\textbf{0.87} \\
\hline
\multirow{4}{*}{\centering Gossipcop} 
    & L & \cellcolor{gossipLight}0.54 & \cellcolor{gossipMedium}0.58 & \cellcolor{gossipLight}0.56 & \cellcolor{gossipMedium}\textbf{0.63} & \cellcolor{gossipMedium}0.62 \\
    & M & \cellcolor{gossipMedium}0.62 & \cellcolor{gossipMedium}0.63 & \cellcolor{gossipMedium}0.58 & \cellcolor{gossipDark}0.68 & \cellcolor{gossipDark}\textbf{0.72} \\
    & G & \cellcolor{gossipDark}0.72 & \cellcolor{gossipDark}0.72 & \cellcolor{gossipMedium}0.64 & \cellcolor{gossipDark}0.71 & \cellcolor{gossipDark}\textbf{0.80} \\
    & L+M+G (MV) & \cellcolor{gossipDark}0.69 & \cellcolor{gossipDark}0.70 & \cellcolor{gossipMedium}0.65 & \cellcolor{gossipDark}0.74 & \cellcolor{gossipDark}\textbf{0.75} \\
\hline
\multirow{4}{*}{\centering Politifact} 
    & L & \cellcolor{politiDark}\textbf{0.88} & \cellcolor{politiDark}0.85 & \cellcolor{politiMedium}0.77 & \cellcolor{politiMedium}0.77 & \cellcolor{politiDark}\textbf{0.88} \\
    & M & \cellcolor{politiMedium}0.73 & \cellcolor{politiMedium}0.77 & \cellcolor{politiMedium}0.77 & \cellcolor{politiLight}0.69 & \cellcolor{politiDark}\textbf{0.86}\\
    & G & \cellcolor{politiDark}0.85 & \cellcolor{politiDark}0.81 & \cellcolor{politiMedium}0.77 & \cellcolor{politiDark}0.81 & \cellcolor{politiDark}\textbf{0.86}\\
    & L+M+G (MV) & \cellcolor{politiDark}0.88 & \cellcolor{politiDark}0.89 & \cellcolor{politiDark}0.82 & \cellcolor{politiDark}0.85 & \cellcolor{politiDark}\textbf{0.91}\\
\bottomrule
\end{tabular}
}

}

\end{table}

\begin{table}[t]
\centering%

\caption{Comparing the performance of MRs with each other using the Paired t-test p-values for AUROC scores. ARI denotes the Average Relative Improvement. n.s.\ =\ not significant. Double red stars (\textcolor{red}{\textbf{**}}) indicate $p < 0.01$.
}
\label{tab:pvalues}
{
\begin{tabular}{lccc}
\toprule
\textbf{Method Comparison} & \textbf{p-value} & \textbf{ARI (\%)} & \textbf{Superior Method} \\
\midrule
PCS vs No MR & $<0.01$\,\textcolor{red}{\textbf{**}} & 8.67 & PCS \\
PCS vs MR1 & $<0.01$\,\textcolor{red}{\textbf{**}} & 10.30 & PCS \\
PCS vs MR2 & $<0.01$\,\textcolor{red}{\textbf{**}} & 31.03 & PCS \\
PCS vs MR3 & $<0.01$\,\textcolor{red}{\textbf{**}} & 9.95 & PCS \\
No MR vs MR1 & 0.35 & 0.03 & n.s. \\
No MR vs MR2 & $<0.01$\,\textcolor{red}{\textbf{**}} & 19.49 & No MR \\
No MR vs MR3 & 0.37 & 0.10 & n.s. \\
MR1 vs MR2 & $<0.01$\,\textcolor{red}{\textbf{**}} & 19.54 & MR1 \\
MR1 vs MR3 & 0.63 & 0.40 & n.s. \\
MR3 vs MR2 & $<0.01$\,\textcolor{red}{\textbf{**}} & 18.82 & MR3 \\
\bottomrule
\end{tabular}

}

\end{table}







\subsection{Impact of Dataset Size on Optimization Stability}

\begin{figure}[t]
  \centering
  \includegraphics[width=1\linewidth]{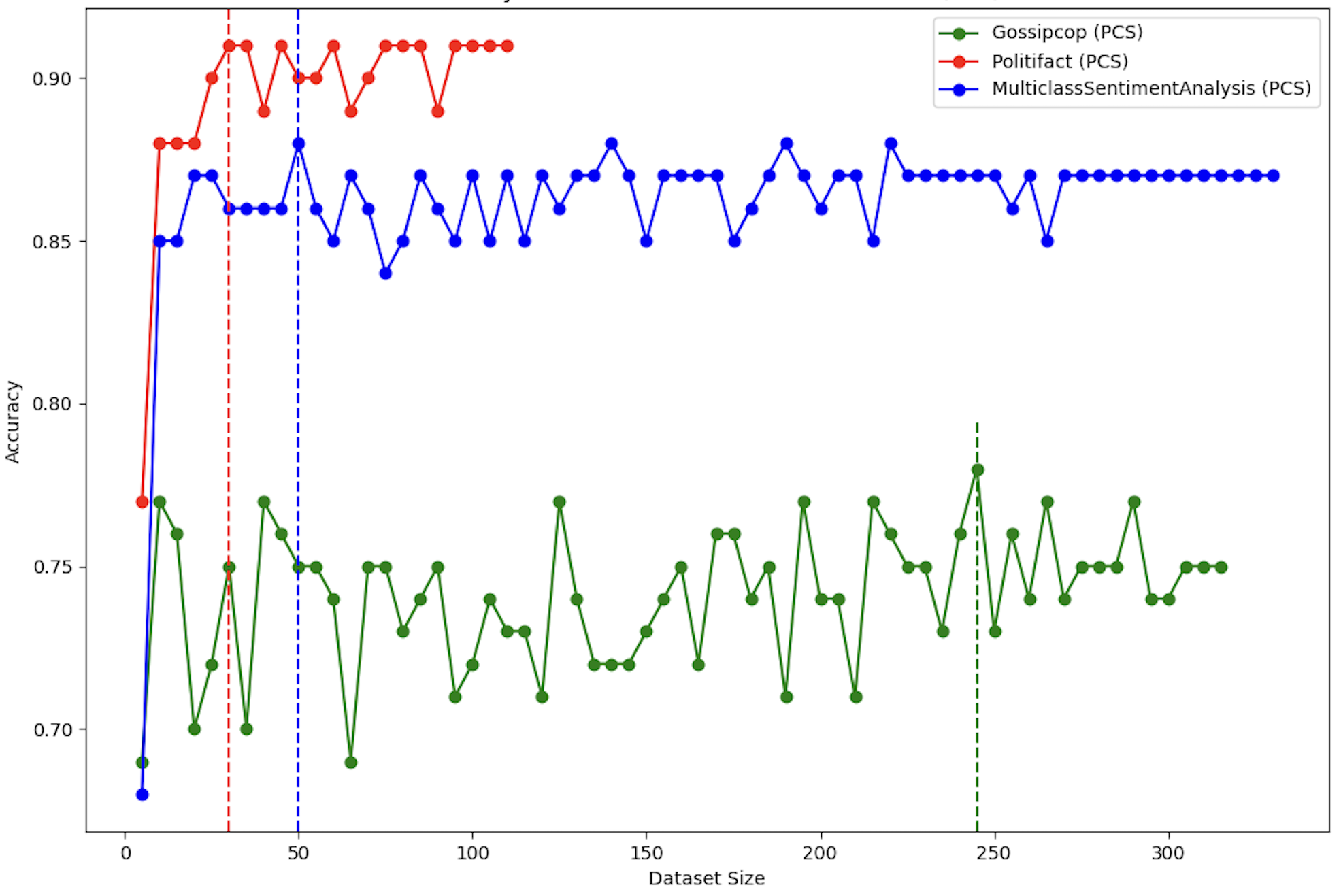} 
  \caption{Effect of optimizing dataset size on AUROC score using the PCS}
    \label{fig:optimizing-datasetsize}
\end{figure}

Figure~\ref{fig:optimizing-datasetsize} shows how dataset size influences hyperparameter optimization (HPO) using Linear Regression (LR) within the PCS framework. The results indicate that convergence rates vary across datasets: \textit{Gossipcop} stabilizes at around 245 samples, \textit{Politifact} at just 30, and \textit{Multiclass Sentiment Analysis} at approximately 50.

These differences suggest that dataset complexity and label consistency affect the amount of data needed for stable weight learning. Notably, PCS remains effective even with limited data, demonstrating its suitability for low-resource settings.


\section{PCS Toolbox}
\label{sec:pcs-annotator}

To facilitate reproducibility and adoption of the PCS framework, we developed an open-source Python library, \texttt{pcs-annotator}
available on PyPI\footnote{\url{https://pypi.org/project/pcs-annotator/}} and GitHub\footnote{\url{https://github.com/salsina/PCS-Python-Library}}. 

The package automates the full PCS pipeline, i.e., from text mutation via Metamorphic Relations (MRs) to multi-LLM annotation, weight optimization, and confidence computation, allowing users to apply PCS to diverse text classification tasks with minimal setup.

\begin{figure}[t]
  \centering
  \includegraphics[width=0.9\linewidth]{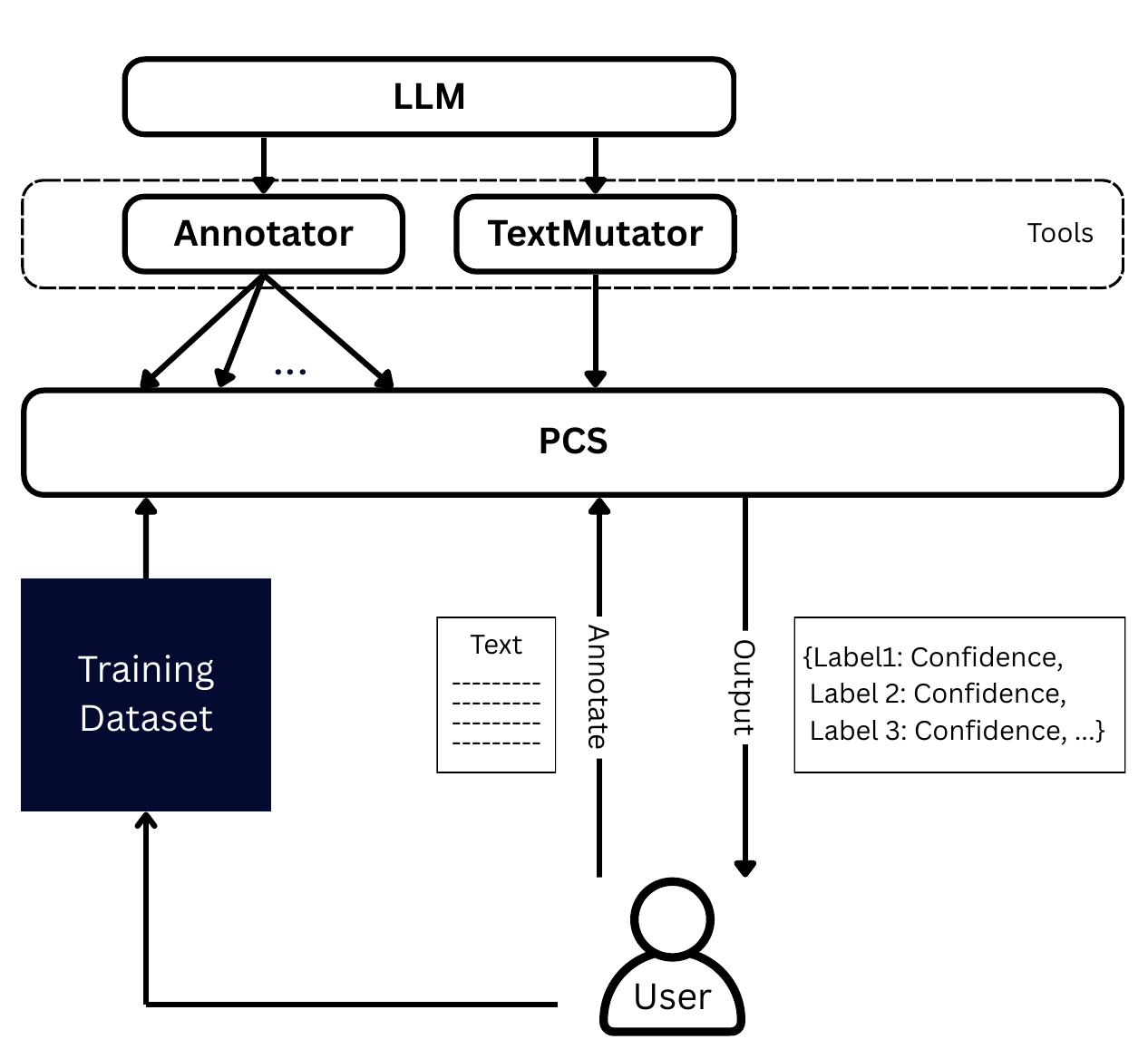}
  \caption{Overview of the \texttt{pcs-annotator} architecture. The system integrates four core components: LLM, Annotator, TextMutator, and PCS modules, orchestrating the data flow from text mutation to confidence computation.}
  \label{fig:pcs-annotator-architecture}
\end{figure}

The library follows a modular and extensible design built upon the LangChain framework \cite{langchain}, enabling integration with multiple LLM providers (e.g., OpenAI, Anthropic, Groq, Hugging Face). It has four main components (Figure \ref{fig:pcs-annotator-architecture}):

\begin{itemize}[leftmargin=10pt]
    \item \textbf{LLM Module:} Initializes and manages model backends through LangChain’s unified interfaces.
    \item \textbf{Annotator Module:} Handles prompt-based annotation and output parsing for classification tasks.
    \item \textbf{TextMutator Module:} Applies predefined or user-defined MRs (e.g., passive/active voice conversion, double negation, synonym substitution) to generate semantically equivalent text variants.
    \item \textbf{PCS Module:} Orchestrates the pipeline, including dataset loading, annotation generation, weight optimization (via Linear Regression or Genetic Algorithms), and PCS computation.
\end{itemize}

A minimal example of using the library is shown below:

\begin{lstlisting}[style=pcs, caption={Example usage of the \texttt{pcs-annotator} library.}]
from pcs_annotator import PCS

pcs = PCS(
    prompt="Label the news as <label>Fake</label> or <label>Real</label>.",
    dataset_path="dataset.csv",
    annotators=["llama3-8b-8192", "mistralai/Mistral-7B-Instruct-v0.3"],
    textmutator="llama-3.1-8b-instant",
    GROQ_API_KEY="your_key",
    generate_annotations=True
)

result = pcs.annotate("Aliens landed in New York yesterday.")
print(result)  # {'Fake': 0.91, 'Real': 0.09}
\end{lstlisting}

The library is easily extensible, supporting new model backends, Metamorphic Relations, or optimization strategies through modular abstractions. Released under MIT open-source license, \texttt{pcs-annotator} promotes transparency, reproducibility, and community-driven research in confidence-based annotation with LLMs.

\section{Threats to Validity}

While the PCS framework demonstrates strong and consistent performance, several factors may affect its generalizability. \textbf{MR Design and Coverage:} PCS currently relies on a fixed set of predefined Metamorphic Relations (MRs). Although these are task-agnostic and semantically grounded, they may not capture the full range of linguistic variation or adversarial scenarios seen in real-world data. More MRs, such as paraphrasing, context expansion, or controlled noise injection, could further improve robustness, provided they are carefully designed to preserve the original meaning. \textbf{Dataset Scope:} Our evaluation covers three classification datasets across binary and multiclass settings, including sentiment analysis and misinformation detection. While these are representative and commonly used, applying PCS to additional tasks such as topic categorization, stance detection, or domain-specific classification would help validate its effectiveness in broader contexts. \textbf{Calibration Dependence:} PCS requires a labeled validation set to learn optimal weights. While it performs well even with small calibration sets, performance may vary depending on the alignment between calibration and test distributions.

\section{Conclusion}

In this work, we proposed the Perceived Confidence Score (PCS), a black-box, perturbation-based framework for evaluating LLM confidence in classification tasks. PCS measures consistency across semantically equivalent inputs and learns task-specific weights over models and transformations to yield robust, interpretable confidence scores.

Evaluated across three datasets and LLMs (LLaMA-3-8B, Mistral-7B, and Gemma-2-9B), PCS improved AUROC by 9.3\% in single-LLM and 5.8\% over majority voting in multi-LLM settings. These gains reflect PCS's ability to capture both linguistic and model-level uncertainty. Beyond improved performance, PCS offers transparency into LLM behavior and supports future extensions like dynamic MR generation and human-in-the-loop workflows.



%

\bibliographystyle{plain}
\bibliography{my-bib}












\newpage

\vfill

\end{document}